\documentclass[conference]{IEEEtran}
\IEEEoverridecommandlockouts
\usepackage{cite}
\usepackage{amsmath,amssymb,amsfonts}
\usepackage{graphicx}
\usepackage{textcomp}
\usepackage{xcolor}
\usepackage{algorithm}
\usepackage{algpseudocode}   % preferred over algorithmic
\usepackage{caption}
\usepackage{url}

\def\BibTeX{{\rm B\kern-.05em{\sc i\kern-.025em b}\kern-.08em
    T\kern-.1667em\lower.7ex\hbox{E}\kern-.125emX}}
\begin{document}

\title{\title{Physics-Guided Counterfactual Explanations for Large-Scale Multivariate Time Series: Application in Scalable and Interpretable SEP Event Prediction}
}

\author{\IEEEauthorblockN{Pranjal Patil}
\IEEEauthorblockA{Dept. of Computer Science \\
Georgia State University\\
Atlanta, GA, USA \\
ppatil8@gsu.edu}
\and
\IEEEauthorblockN{Anli Ji}
\IEEEauthorblockA{Dept. of Computer Science\\
California State University\\
Fullerton, CA, USA \\
anli.ji@fullerton.edu}
\and
\IEEEauthorblockN{Berkay Aydin}
\IEEEauthorblockA{Dept. of Computer Science \\
Georgia State University\\
Atlanta, GA, USA \\
baydin2@gsu.edu}
}

\maketitle

\begin{abstract}
Accurate prediction of solar energetic particle events is vital for safeguarding satellites, astronauts, and space-based infrastructure \cite{Fogtman_2023}. Modern space weather monitoring generates massive volumes of high-frequency, multivariate time series (MVTS) data from sources such as the Geostationary Operational Environmental Satellites (GOES). Machine learning (ML) models trained on this data show strong predictive power, but most existing methods overlook domain-specific feasibility constraints. Counterfactual explanations have emerged as a key tool for improving model interpretability, yet existing approaches rarely enforce physical plausibility. This work introduces a Physics-Guided Counterfactual Explanation framework, a novel method for generating counterfactual explanations in time series classification tasks that remain consistent with underlying physical principles. Applied to solar energetic particles (SEP) forecasting, this framework achieves over 80\% reduction in Dynamic Time Warping (DTW) distance increasing the proximity, produces counterfactual explanations with higher sparsity, and reduces runtime by nearly 50\% compared to state-of-the-art baselines such as DiCE\cite{Mothilal2020}. Beyond numerical improvements, this framework ensures that generated counterfactual explanations are physically plausible and actionable in scientific domains. In summary, the framework generates counterfactual explanations that are both valid and physically consistent, while laying the foundation for scalable counterfactual generation in big data environments.
\end{abstract}


\section{Introduction}
Solar Energetic Particle (SEP) events, originating from solar flares and coronal mass ejections, are bursts of high-energy protons and ions that pose significant risks to both space-borne and ground-based technologies, including satellite electronics, astronaut health, aviation communication, and terrestrial power systems. As one of the most consequential solar activities, SEP events drive interplanetary disturbances and exhibit strong coupling with geomagnetic variations, underscoring the importance of accurate and timely prediction in space weather forecasting \cite{Vainio2009,Fogtman_2023}. 

Over the past decades, continuous monitoring missions such as the Geostationary Operational Environmental Satellites (GOES) have produced vast amounts of multivariate time-series (MVTS) data. These datasets record proton flux intensities across multiple energy channels, sampled at high temporal resolution over decades of solar cycles. The scale, heterogeneity, and temporal richness of these observations make SEP forecasting a canonical big data problem: models must integrate large volumes of sequential, multi-channel measurements while capturing nonlinear and nonstationary dynamics. Recent machine learning (ML) approaches have demonstrated strong predictive potential, complementing traditional physics-based models \cite{10431590,10386908}. 

Despite promising performance, current ML approaches remain largely black-box. Lack of interpretability constrains scientific insight, undermines trust in operational deployment, and limits adoption in high-stakes environments. In the big data context, interpretability is especially critical: as models consume ever-larger MVTS streams, their outputs must remain explainable to scientists and decision-makers alike. Existing explanation frameworks, such as counterfactual explanations, show promise in tabular domains but rarely account for the structured, temporal, and physics-constrained nature of scientific sensor data, limiting their utility in SEP forecasting. 

To address these limitations, we introduce a Physics-Guided Counterfactual Explanation framework for multivariate time-series data. This approach generates counterfactual explanations that not only alter model predictions (for example, ``non-Solar Energetic Particle Event’’ to ``Solar Energetic Particle Event’’) but also remain physically plausible by preserving flux continuity, cross-energy correlations, and onset delays across proton channels. A novel reconstruction component maps counterfactual perturbations back into the original time series, producing temporally consistent and scientifically meaningful explanations that can be directly interpreted by domain experts. When applied to Solar Energetic Particle forecasting, the framework produces counterfactual explanations with lower Dynamic Time Warping distance, higher sparsity, and reduced runtime compared to state-of-the-art baselines such as the DiCE\cite{Mothilal2020} framework. Importantly, this approach generates explanations that are both physically plausible and consistent with domain knowledge, bridging the gap between data-driven forecasts and scientific interpretability while providing a foundation for generation of scalable counterfactual explanations in big data environments.

Specifically, the contributions of this work include:
\begin{enumerate}
    \item Adaptation of DiCE-like counterfactual methods to large-scale MVTS data from GOES, ensuring scalability to decades of continuous proton flux measurements.
    \item Integration of domain knowledge from solar physics, such as flux ratios, onset delays, and smooth temporal transitions into the counterfactual optimization process, guaranteeing scientific plausibility.
    \item Demonstration that physics-guided counterfactual explanations enhance the interpretability and trustworthiness of ML forecasts for SEP events.
    \item Introduction of a reconstruction framework that translates counterfactual perturbations into full, temporally consistent time series, enabling actionable insights.
    \item Empirical evaluation on labeled GOES datasets, showing superior performance over unconstrained baselines in generating realistic counterfactual explanations with improvements in faithfulness, sparsity, and interpretability.
\end{enumerate}

\section{Related Work}
\label{section:Problem_formulation}
Counterfactual explanations are increasingly used to make machine learning models more interpretable. While initial research focused on tabular data, recent studies have extended counterfactual explanations to sequential and high-volume datasets. Theissler et al. \cite{Theissler2022} review methods for generating counterfactual explanations in time-series data. Some approaches, like Native Guide \cite{Delaney2021}, rely on real examples to improve plausibility, while optimization-based methods such as $\tau$RT and $\tau$IRT \cite{Karlsson2019} generate counterfactual explanations through constrained searches over time-series classifiers. Other techniques, like CoMTE \cite{Ates2021}, reuse authentic training segments for realism, and contrastive methods \cite{labaien2020contrastive} use deep learning to create meaningful negative examples. Together, these approaches address realism, diversity, and interpretability in time-series counterfactual explanations.

General-purpose frameworks such as DiCE\cite{Mothilal2020} remain popular due to their strong performance on proximity, diversity, feasibility, and computational efficiency. However, off-the-shelf DiCE and similar algorithms do not provide mechanisms to embed domain knowledge or enforce physics-based constraints, limiting their suitability for scientific and operational domains. In SEP forecasting, unconstrained counterfactual explanations can violate physical consistency across proton flux channels and thus be scientifically misleading.

Beyond algorithmic design, practical deployment in operational settings raises explicit big data concerns: GOES and similar observational platforms produce high-frequency, multi-channel streams with substantial volume, velocity, and variety; these datasets require scalable preprocessing, robust handling of missing/erroneous measurements (data quality), and efficient model/explainer evaluation in batch or streaming modes. Prior time-series counterfactual explanation work typically omits these system-level considerations, resulting in methods that may not scale or maintain plausibility under real-world data quality issues.

In contrast to prior work that
(a) focuses on tabular inputs (e.g. DiCE)
(b) reuses real time-series segments without enforcing domain constraints (e.g. CoMTE\cite{Ates2021})
(c) performs unconstrained temporal optimization (e.g. $\tau$RT/$\tau$IRT \cite{Karlsson2019}), our approach combines three complementary innovations: 
(1) a window-aware encoding for MVTS that groups semantically related features to respect temporal context and reduce search dimensionality; 
(2) constraint-aware genetic operators and penalty-driven objectives that enforce physical ordering, per-channel range bounds, and temporal smoothness during candidate generation; and 
(3) a practical local-global reconstruction procedure that maps per-window counterfactual explanations back to continuous time series while preserving temporal coherence. Importantly, these design choices are implemented with attention to big-data constraints (parallel population evaluation, batch processing, and robustness to missing values), enabling counterfactual explanations that are both faithful to the predictive model and physically plausible in operational SEP forecasting scenarios.

\section{Methodology}
\subsection{Problem Formulation}
\label{subsection:Problem_formulation}
Given a trained classifier \( f: \mathcal{X} \rightarrow \mathcal{Y} \) that maps an input multivariate time-series instance \( \mathbf{X} \in \mathcal{X} \) to its predicted class label \( y \in \mathcal{Y} \) (e.g., SEP or non-SEP event), the objective is to generate a counterfactual instance \( \mathbf{X}' \) that minimally perturbs the original input while flipping the model prediction, such that \( f(\mathbf{X}') \neq f(\mathbf{X}) \).  These counterfactual explanations must:
1) Flip the model's prediction (e.g., from SEP event to Non-SEP Event), 2) Make minimal, physically plausible adjustments to the features of input instance \(\mathbf{X}\), 3) Satisfy domain-specific physical constraints derived from solar physics.

The multivariate input time series \( \mathbf{X} \) can be represented as a feature matrix:
\[
\mathbf{X} = 
\begin{bmatrix}
x_{1}[w_1] & x_{1}[w_2] & \cdots & x_{1}[w_W] \\
x_{2}[w_1] & x_{2}[w_2] & \cdots & x_{2}[w_W] \\
\vdots & \vdots & \ddots & \vdots \\
x_{m}[w_1] & x_{m}[w_2] & \cdots & x_{m}[w_W]
\end{bmatrix}
\in \mathbb{R}^{m \times W},
\]
where each row corresponds to a physically related proton-flux channel (e.g., \( P3, P5, P7 \)), and each column corresponds to a temporal window \( w \in \{1, \dots, W\} \). 
Each element \( x_i[w_j] \) thus represents the feature value of channel \( i \) within window \( w_j \), capturing a statistical descriptor such as the mean proton flux intensity.

For compatibility with standard counterfactual generation frameworks, the feature matrix \( \mathbf{X} \) is vectorized into a single input vector,
\[
\mathbf{x} = \operatorname{vec}(\mathbf{X}) = [x_{1}[w_1], x_{1}[w_2], \dots, x_{1}[w_W], x_{2}[w_1], \dots, 
\]
\[
x_{m}[w_W]] \in \mathbb{R}^{m \times W},
\]
where \( \mathbb{R}^{m \times W} \) denotes the total number of features across all channels and time windows. This vectorized form \( \mathbf{x} \) serves as the model input for \( f(\mathbf{x}) \) and is used directly in the optimization objective during counterfactual generation. The corresponding counterfactual instance is represented as \( \mathbf{X}' \in \mathbb{R}^{m \times W}  \), with elements \( x_i'[w_j] \) that satisfy domain-specific physical constraints derived from solar energetic particle (SEP) dynamics as discussed below:

\begin{enumerate}
    \item \text{Feature Ordering (within each time window):}
    For each temporal window \( w_j \), let 
    \[
    \mathcal{O}_{w_j} = \{x_{1}[w_j], x_{2}[w_j], \dots, x_{m}[w_j]\}
    \]
    denote the ordered set of flux-channel features across energy levels. 
    Empirically, lower-energy channels exhibit higher proton-flux intensities, resulting in a physical ordering such that
    \[
    x_{1}[w_j] \geq x_{2}[w_j] \geq \dots \geq x_{m}[w_j],
    \]
    e.g. with respect to the channel hierarchy \( P3 > P5 > P7 \) (where \( P3 > 10~\mathrm{MeV},\, P5 > 50~\mathrm{MeV},\, P7 > 100~\mathrm{MeV} \)). 
    The counterfactual matrix \( \mathbf{X}' \) must preserve this ordering in every window:
    \[
    x_{1}'[w_j] \geq x_{2}'[w_j] \geq \dots \geq x_{m}'[w_j], \quad \forall\, w_j \in \{1, \dots, W\}.
    \]

    \item \text{Feature Range Constraint (per channel per window):}
    For each element \( x_i[w_j] \) in \( \mathbf{X} \), the counterfactual value \( x_i'[w_j] \) must remain within its empirically observed range:
    \[
    x_i^{\min}[w_j] \leq x_i'[w_j] \leq x_i^{\max}[w_j],
    \]
    where \( x_i^{\min}[w_j] \) and \( x_i^{\max}[w_j] \) are derived from the historical distribution of proton fluxes for channel \( i \) in window \( w_j \). 
    This constraint prevents counterfactual explanations from producing physically implausible values.

    \item \text{Temporal Consistency (across adjacent windows):}
    The time-evolution of flux values must remain smooth across consecutive windows. 
    For each channel \( i \), transitions between adjacent time windows \( w_j \) and \( w_{j+1} \) should satisfy:
    \[
    |x_i'[w_{j+1}] - x_i'[w_j]| \leq \delta_i,
    \]
    where \( \delta_i \) is a small, channel-specific tolerance controlling temporal smoothness. 
    This constraint penalizes abrupt, non-physical fluctuations in \( \mathbf{X}' \) and ensures realistic temporal behavior of the reconstructed counterfactual series.
\end{enumerate}

Unlike traditional tabular counterfactual frameworks, the proposed formulation explicitly preserves the spatiotemporal structure of the multivariate time series \( \mathbf{X} \) through:

\begin{enumerate}
    \item \text{Multivariate Interactions:}
    Coupled constraints across proton-flux channels within each window \( w_j \), maintaining physically meaningful correlations among flux measurements.

    \item \text{Temporal Dependencies:}
    Penalty terms that regularize discontinuities across consecutive windows to enforce smooth temporal evolution and consistent event progression.

    \item \text{Window-Aware Optimization:}
    Constraint enforcement is applied per window, using channel groupings defined by their temporal index \( w_j \). 
    This structure enables optimization to preserve intra-window ordering and inter-window continuity simultaneously.
\end{enumerate}

Given these physically grounded constraints, the generation of the counterfactual instance \( \mathbf{X}' \) (or equivalently its vectorized form \( \mathbf{x}' = \operatorname{vec}(\mathbf{X}') \)) is formulated as a constrained optimization problem that balances proximity, validity, and physical plausibility. 
The objective function is defined as:

\[
\mathbf{x}^{\prime} = 
\arg\min_{\mathbf{x}^{\prime}} 
\underbrace{\|\mathbf{x}^{\prime} - \mathbf{x}\|_1}_{\text{Proximity}} 
+ 
\lambda_1 \underbrace{\mathcal{L}_\text{valid}(f(\mathbf{x}'))}_{\text{Validity}} 
+ 
\lambda_2 \underbrace{\mathcal{P}_\text{order}(\mathbf{x}^{\prime})}_{\text{Ordering}} 
\]

\[
+ 
\lambda_3 \underbrace{\mathcal{P}_\text{range}(\mathbf{x}^{\prime})}_{\text{Range}}
\]
where each term corresponds to a specific objective in the counterfactual generation process:

\begin{itemize}
    \item \textbf{Proximity:} 
    The term \( \|\mathbf{x}' - \mathbf{x}\|_1 \) enforces minimal deviation between the counterfactual instance and the original input, ensuring that the generated explanation modifies as few features as possible across all channel–window combinations \( (i, w_j) \). 

    \item \textbf{Validity:} 
    The loss term \( \mathcal{L}_\text{valid}(f(\mathbf{x}')) \) penalizes counterfactual explanations that fail to flip the model’s prediction, ensuring that \( f(\mathbf{x}') \neq f(\mathbf{x}) \). 
    This guarantees that the generated instance belongs to the target decision boundary of the classifier \( f \).

    \item \textbf{Ordering Penalty:} 
    The term \( \mathcal{P}_\text{order}(\mathbf{x}') \) quantifies violations of the physical ordering constraint within each window \( w_j \), penalizing instances where
    \[
    x_{1}'[w_j] < x_{2}'[w_j] \ \text{or} \ x_{2}'[w_j] < x_{3}'[w_j],
    \]
    thereby maintaining the expected monotonic relationship among proton-flux channels (\( P3 > P5 > P7 \)).

    \item \textbf{Range Penalty:} 
    The function \( \mathcal{P}_\text{range}(\mathbf{x}') \) penalizes deviations of \( x_i'[w_j] \) beyond their empirically observed bounds:
    \[
    x_i^{\min}[w_j] \le x_i'[w_j] \le x_i^{\max}[w_j],
    \]
    preserving physically plausible flux magnitudes across channels and windows.
\end{itemize}

The weighting coefficients \( \lambda_1, \lambda_2, \lambda_3 \geq 0 \) control the trade-offs among prediction validity, physical ordering, and range adherence. 

This formulation ensures generated counterfactual explanations are both actionable (respect physical laws) and interpretable (preserve MVTS structure), addressing key limitations of generic counterfactual methods in solar physics applications. We extended the DiCE\cite{Mothilal2020} framework by implementing a custom ConstrainedDiceGenetic class, which enforces the above domain-specific range and temporal ordering constraints through penalty terms during candidate generation.

\subsection{Dataset Details}
\label{subsection:dataset_details}
In this study, the publicly available Geostationary Solar Energetic Particle (GSEP) event dataset published on the Harvard Dataverse\cite{GSEP2022} was used. This dataset consolidates 341 SEP events spanning three complete solar cycles (1986–2017) by integrating multiple catalogs (PSEP, CDAW-SEP, and NOAA-SEP). It provides minute-level proton flux measurements from the GOES-13 and GOES-15 satellites across several energy channels, along with heterogeneous metadata describing associated solar flares, CMEs, active regions, and radio bursts, recorded at a 5-minute cadence. For the present analysis, three key proton flux channels- P3 ($>10$~MeV), P5 ($>50$~MeV), and P7 ($>100$~MeV) were utilized, as they are critical for SEP detection and characterization. Each instance was labeled to indicate SEP occurrence, enabling reliable supervised learning. To capture temporal dynamics at multiple scales, a sliding-window approach with window lengths ranging from 6 to 360 minutes was applied, and mean flux values per channel within each interval were extracted as feature vectors for modeling. The scale and heterogeneity of the GSEP dataset exemplify its BigData characteristics-high volume (decades of continuous observations), high variety (flux profiles and event-level metadata), high velocity (near-real-time cadence), and high veracity (cross-catalog reconciliation). These properties establish GSEP as a large-scale, multidimensional benchmark ideally suited for advanced machine learning and BigData-driven space weather forecasting.

\subsection{Evaluation of Candidate Models and Baseline Selection }
\label{subsection:evaluation_of_candidates}
Convolution-based approaches such as ROCKET \cite{rocket} provide strong predictive performance in time series classification due to their ability to capture local temporal dynamics at scale. Their feature space is also compatible with counterfactual generation frameworks such as DiCE, which can perturb derived statistics (e.g., maximum, average, or kernel activations) to produce alternative outcomes. However, these convolution-derived features are not directly interpretable in the raw time domain, making the reconstruction of counterfactual instances highly non-trivial. Since multiple raw series can yield the same feature values, mapping a perturbed feature vector back into a feasible time series remains ill-posed, which in turn limits visualization and interpretability.

In contrast, shapelet-based models such as SAX\cite{Lin2003} or adaptive discriminative shapelets \cite{Li2022} provide a more direct link between feature representation and raw temporal patterns. Counterfactual explanations in this setting can be interpreted as modifications to specific subsequences, which are inherently more meaningful and visually intuitive. This local interpretability enables highlighting regions of the series that drive the classification decision, facilitating counterfactual explanation. Nevertheless, shapelet discovery is computationally expensive and sensitive to alignment and noise, and counterfactual explanations are constrained by the availability of suitable subsequences in the dataset.

Overall, both model families face challenges when integrated with DiCE for counterfactual generation in time series. Convolution-based methods offer efficiency and predictive power but suffer from poor reconstruction feasibility, whereas shapelet-based methods provide higher interpretability but limited flexibility. In both cases, generating realistic counterfactual explanations in the raw time domain remains a fundamental difficulty, underscoring the need for hybrid strategies that combine feature-based perturbation with generative reconstruction mechanisms to ensure feasibility and improve visualization.

We adopt the DiCE explainer with its genetic algorithm backend, which efficiently searches for diverse and feasible counterfactual explanations without requiring gradient information, making it well-suited for tree-based models such as Random Forests. The Random Forest Classifier was selected as our baseline due to its inherent interpretability, robustness against overfitting, and strong performance in handling high-dimensional feature spaces. Furthermore, preliminary experiments \cite{Ji2024}, \cite{10386908}, \cite{10431590} demonstrated that Random Forest consistently outperformed other ensemble methods, reinforcing its suitability for this task. We restrict our experimental evaluation to Random Forests as the baseline, since convolution- and shapelet-based models, while effective for classification, pose challenges for counterfactual generation and reconstruction, as discussed earlier. A detailed empirical comparison is outside the scope of this work.

\subsection{Explainer Model to Generate DiCE Based Counterfactual Explanations with Physical Constraints }
\label{subsection:DiCE_explainer_model}
We use the DiCE explainer with the genetic algorithm and Random Forest Classifier as mentioned in section  This scikit-learn back-end configuration ensures compatibility with DiCE’s tabular data format and enables the use of evolutionary optimization to generate realistic, actionable, and physically plausible counterfactual explanations as proposed in Subsection [\ref{subsection:Problem_formulation}]. The process evolves a population of candidate solutions from the original SEP event instance, optimizing for proximity, validity, and domain-specific physical constraints as described in Algorithm [\ref{alg:genetic_cfe_generation}].

\begin{algorithm}[t]
\caption{\textbf{Constrained Counterfactual Generation with Genetic DiCE}}
\label{alg:genetic_cfe_generation}

\textbf{0: Input:} Training data $df_{\text{train}}$, classifier $f$, permitted ranges $\mathcal{R}$, ordering penalty $\lambda_{\text{order}}$ \\
\textbf{0: Output:} Counterfactual explanation $x'_{\text{smooth}}$

\textbf{1: Feature Extraction:}\\
$C \gets df_{\text{train}} \setminus \{\text{Label}\}; \;
W \gets \{ w \mid w = [t_i : t_j] \}$

\textbf{2: Initialize DiCE:}  \\
\hspace{1.2em} $data \gets dice\_ml.Data(df\_cfe, C, Label)$ \\ 
\hspace{1.2em} $model \gets dice\_ml.Model(f,'sklearn','classifier')$

\textbf{3: Define Constraints:} \\ 
\hspace{1.2em} $P_{\text{order}} = \sum_{w \in W} \mathbb{I}(x'_i[w] \leq x'_j[w] \lor x'_j[w] \leq x'_k[w])$  
\hspace{1.2em} $P_{\text{range}} = \sum_{x_j \in C} \mathbb{I}(x'_j \notin [x^{\min}_j, x^{\max}_j])$

\textbf{4: Initialize Constrained Explainer:} \\ 
\hspace{1.2em} $exp \gets ConstrainedDiceGenetic(data, model, \mathcal{R}, \lambda_{\text{order}})$

\textbf{5: Generate Counterfactual:}  \\
\hspace{1.2em} $x' = \arg\min_{x'} \|x' - x\|_1 + \lambda_1 L_{\text{valid}} + \lambda_2 P_{\text{order}} + \lambda_3 P_{\text{range}}$  
\hspace{1.2em} using tournament selection, blend crossover, and constrained mutation

\textbf{6: Temporal Smoothing:} $x'_{\text{smooth}} \gets rolling\_mean(x')$

\textbf{7: Return:} $x'_{\text{smooth}}$

\end{algorithm}

Although our approach builds on the existing DiCE framework, its extension to MVTS data in the solar physics domain represents a non-trivial methodological adaptation. Conventional counterfactual explanation methods, including DiCE, are designed for static tabular data and cannot directly accommodate temporal dependencies or domain-specific constraints. To address this gap, we developed a custom ConstrainedDiceGenetic module that incorporates range bounds, temporal smoothness across sliding windows, and feature-ordering relationships derived from physical laws. This design ensures that generated counterfactual explanations remain both data-plausible and scientifically interpretable. To the best of our knowledge, this is the first application of counterfactual explanations to MVTS data in solar physics, thereby demonstrating how generic XAI frameworks can be adapted to support high-dimensional, domain-constrained scientific data. Beyond this specific use case, our formulation illustrates a pathway for extending counterfactual explanation methods to other MVTS settings in large-scale data science applications.

\subsection{Reconstruction of Timeseries Using Local-Global Offsets for the Counterfactual Explanation of a SEP timeseries}
\label{subsection:timeseries_reconstruction}

Most counterfactual explanation methods, including widely used frameworks such as DiCE, produce explanations in the form of modified feature vectors. While this representation is effective for tabular data, it is insufficient for sequential domains such as SEP event prediction, where the model inputs are time series. In these cases, a counterfactual expressed only as perturbed feature values (e.g., local averages) does not directly reveal how the entire temporal signal would appear if those conditions were realized. This creates a gap between counterfactual reasoning and domain interpretability, since domain experts cannot assess the physical plausibility of isolated feature changes without observing their effect on the overall time series.

\begin{algorithm}[t]
\caption{\textbf{Reconstruction of Time-Series for Counterfactual Explanations}}
\label{alg:generic_timeseries}

\textbf{0: Input:} Time-series dataset $D$, counterfactual instance $cfe$, target feature $f$, slice intervals $\mathcal{S}$, observation window $[t_{start}, t_{end}]$ \\  
\textbf{0: Output:} Perturbed series $Y_{\text{pert}}$, original series $Y_{\text{orig}}$, value range $[y_{\min}, y_{\max}]$  

\textbf{1: Initialization:} \\  
\hspace{1.2em} Select observation window $D_{\text{obs}} \subseteq D$ within $[t_{start}, t_{end}]$ \\  
\hspace{1.2em} Initialize accumulation array $A_{\text{accum}} \gets 0$, counter array $C_{\text{count}} \gets 0$  

\textbf{2: Iterative Adjustment:} \\  
For each slice $[s, e] \in \mathcal{S}$:\\{  
\hspace{1.2em} Extract slice values $F \gets D_{\text{obs}}[f, s:e]$ \\  
\hspace{1.2em} Retrieve counterfactual mean $\hat{F}$ from $cfe$ for the same slice \\  
\hspace{1.2em} Compute adjustment $\Delta \gets \hat{F} - \text{mean}(F)$ \\  
\hspace{1.2em} Update $A_{\text{accum}}$ and $C_{\text{count}}$ with adjusted values  
}  

\textbf{3: Output Construction:} \\  
\hspace{1.2em} Derive final offsets $O_{\text{final}} \gets A_{\text{accum}} / C_{\text{count}}$ \\  
\hspace{1.2em} Construct perturbed series $Y_{\text{pert}} \gets Y_{\text{orig}} + O_{\text{final}}$ \\  
\hspace{1.2em} Compute global min–max range $[y_{\min}, y_{\max}]$ across $Y_{\text{orig}}, O_{\text{final}}, Y_{\text{pert}}$  

\textbf{4: Return:} $Y_{\text{pert}}, Y_{\text{orig}}, [y_{\min}, y_{\max}]$

\end{algorithm}

\begin{figure*}[h]
    \centering
    \includegraphics[width=1\linewidth]{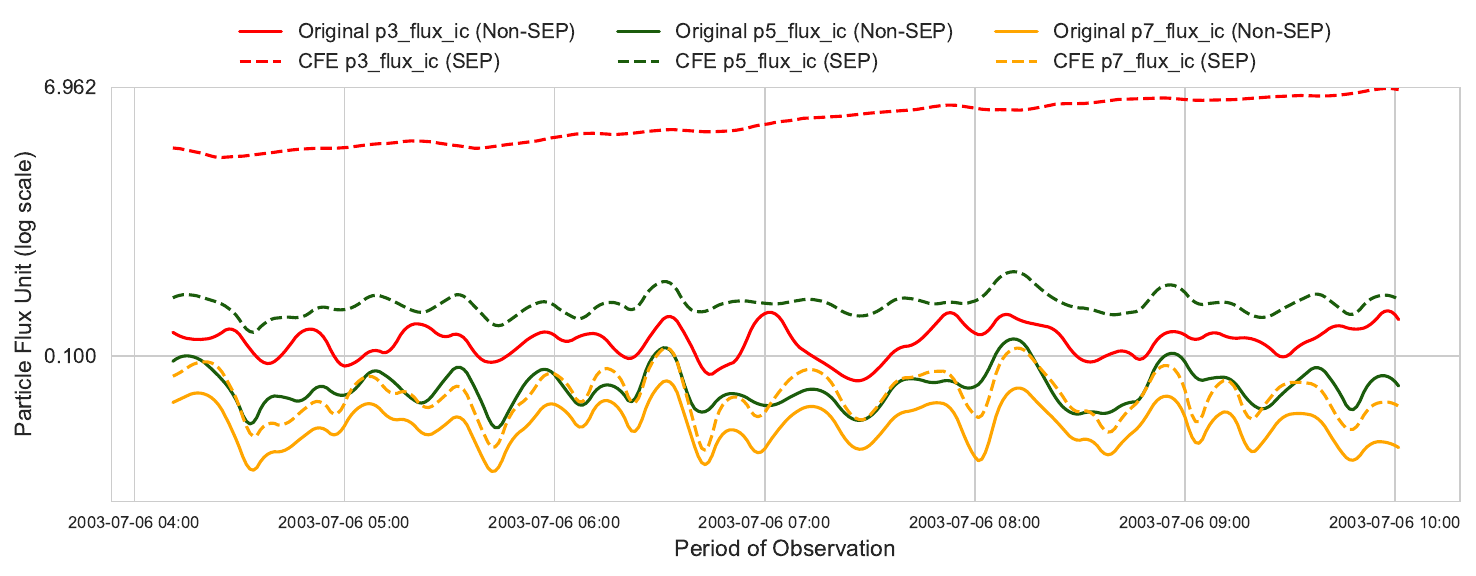}
    \caption{GOES-SEM Solar Proton Flux and their Counterfactual Explanation (CFE) with Blended Overlapping Windows}
    \label{fig:CFE_reconstruction}
\end{figure*}

\begin{figure*}[h]
    \centering
    \includegraphics[width=\textwidth]{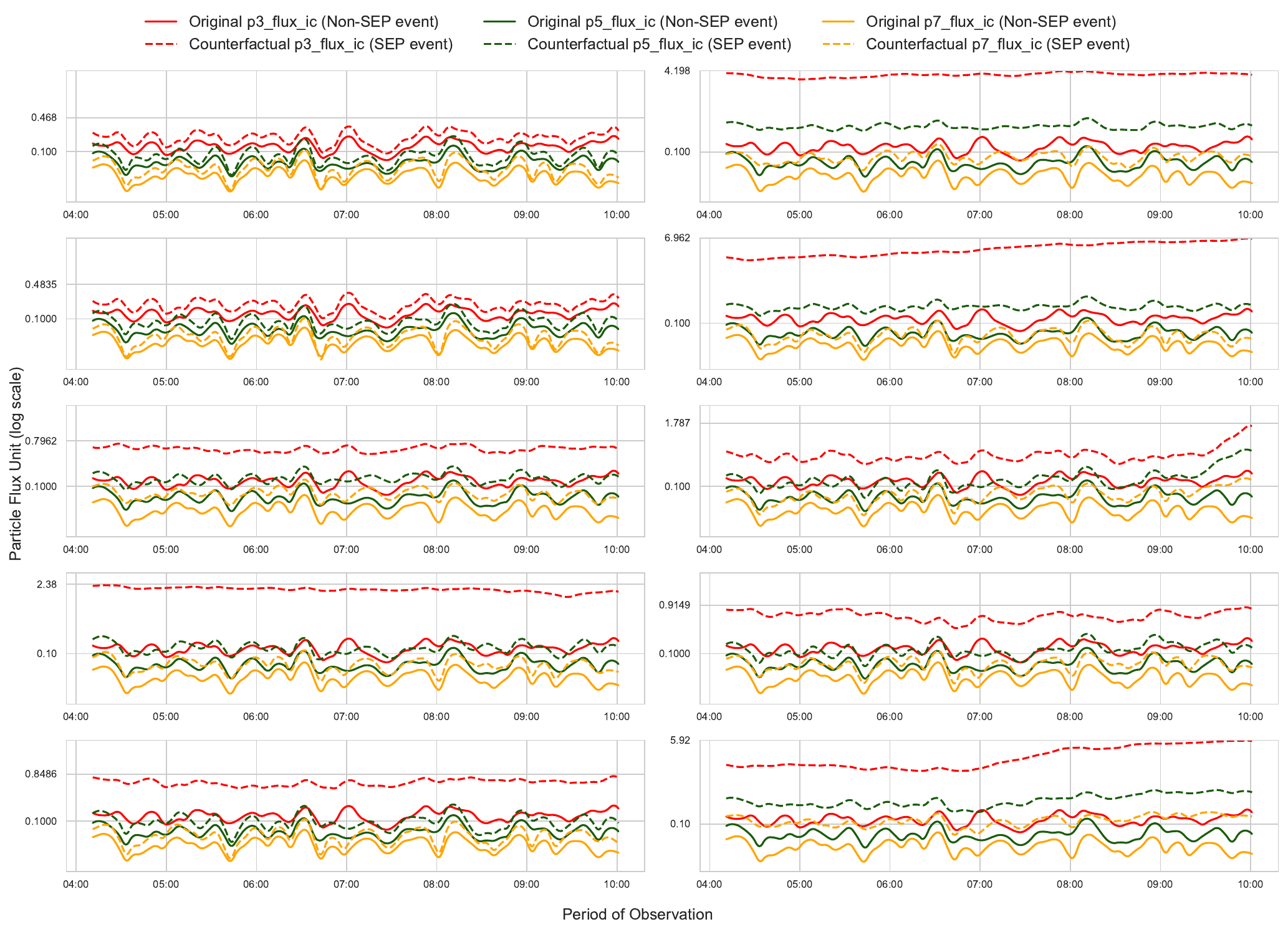}
    \caption{Illustration of diverse counterfactual reconstructions for a single event. The original input corresponds to a non-SEP event, while the generated counterfactual explanations reflect alternative scenarios classified as SEP events.}
    \label{fig:multiple_CFE}
\end{figure*}

To bridge this gap, our approach introduces the reconstruction process described in Algorithm [\ref{alg:generic_timeseries}], which maps perturbed counterfactual values back into a physically realizable time series. The process begins by extracting the relevant segment of the original time series for a given flux type and time window. For each window (or slice), the local mean from the counterfactual explanation is matched by adjusting all data points within the segment, and overlapping adjustments are averaged to ensure smooth transitions. This results in a perturbed series that preserves both the temporal structure and physical plausibility of the original data. By transforming counterfactual explanations from abstract vector modifications into full signal reconstructions, this method enables direct visualization and evaluation of sequential counterfactual explanations. A sample SEP event timeseries and its counterfactual explanation has been visualized in Fig.\ref{fig:CFE_reconstruction} and Fig. \ref{fig:multiple_CFE}.

The ability to generate realistic, temporally consistent signals from counterfactual explanations represents a novel contribution to counterfactual reasoning in time-series domains. Unlike prior work that stops at static feature perturbations, our framework enables explanations to be inspected in the same format as the original scientific data, facilitating interpretability, trust, and domain validation. This work not only advances explainability in SEP event prediction but also generalizes to other sequential data applications such as biomedical signals, financial forecasting, and climate modeling.

\section{Experiments}
\label{sec:experiments}
\subsection{Experiment Design}
\label{subsection:exp_design}
The experimental workflow consisted of three primary stages: first, a Random Forest Classifier was developed for the prediction task, with its hyperparameters optimized through grid search and stratified 5-fold cross-validation to maximize the F1 score for the positive class. Following model training, counterfactual explanations were generated using a customized explainer described in Section [\ref{subsection:DiCE_explainer_model}] based on the DiCE framework. The explainer’s hyperparameters were carefully re-evaluated to explore trade-offs in the quality, plausibility and interpretability of model decisions based on the evaluation metrics from multiple research in \cite{Mazzine2021},\cite{Mothilal2020},\cite{Looveren2019},\cite{Laugel2020},\cite{2007_Springer}.The third phase included fidelity\cite{Velmurugan2021} evaluation of the generated counterfactual explanations.

\subsection{Evaluation Metrics}
\label{sec:evaluation_metric}
Random Forest Classifier training and testing alongwith the Fidelity evaluation included accuracy, True Skill Statistic (TSS) and Heidke Skill Score (HSS). The counterfactual explantions \(\mathbf{x}^{\prime}\) for an original input instance \(\mathbf{x}\) were evaluated based on proximity using Dynamic Time Warping (DTW)\cite{2007_Springer}, sparsity and diversity as mentioned in \cite{Mothilal2020}.

\begin{equation}
\text{DTW}(x,x') = \min_{\text{warping path } W} \sum_{(i,j) \in W} d(x_i, x'_j)
\end{equation}
\begin{equation}
\text{Sparsity}(x, x') = \sum_{i=1}^d \mathbb{I}[x_i \neq x'_i]
\end{equation}
\begin{equation}
\text{Diversity}_{\text{mean}} = \frac{2}{n(n-1)} \sum_{i=1}^{n-1} \sum_{j=i+1}^{n} \|x'_i - x'_j\|_1
\end{equation}

where,
\begin{itemize}
    \item $x = [x_1, \dots, x_d]$, $x' = [x'_1, \dots, x'_d]$: Original and counterfactual feature vectors, used for DTW, sparsity, and diversity computations.
    \item $n$: Number of generated counterfactual explanations for diversity computation.
    \item $\mathbb{I}$: Indicator function, equal to 1 if the feature value changes, otherwise 0.
    \item $x'_i, x'_j$: Two different counterfactual explanations compared for diversity.
\end{itemize}

\subsection{Experiment 1: Physics-guided constraint optimization for explainer model}
\label{exp:exp1}
\begin{figure*}[!h]
    \centering
    \includegraphics[width=0.89\textwidth]{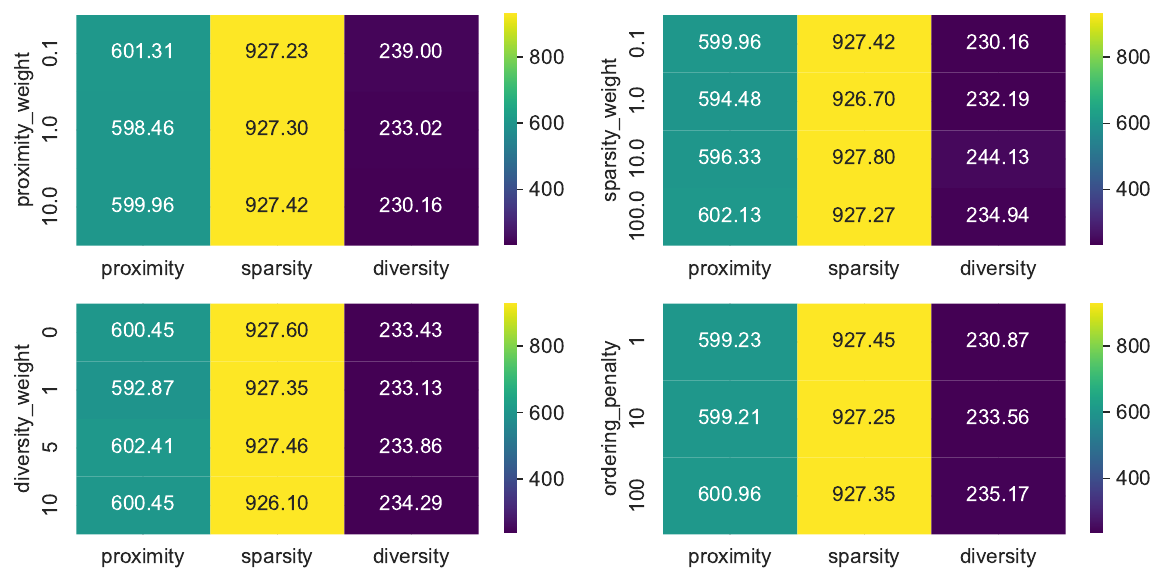}
    \caption{Effect of Hyperparameter Settings on Generation of Counterfactual Explanations' Quality Metrics}
    \label{fig:exp1_fig}
\end{figure*}

The Physics-guided constraint optimization experiment was conducted using a grid search over key hyperparameters: \textit{proximity\_weight}, \textit{sparsity\_weight}, \textit{diversity\_weight} extended from DiCE, and along with physics-guided \textit{ordering\_penalty} introduced in Section [\ref{subsection:Problem_formulation}], to examine their influence on counterfactual quality. For each hyperparameter configuration, counterfactual explanations were evaluated using three core metrics: proximity, sparsity, and diversity, as described in Section [\ref{sec:evaluation_metric}]. This approach enabled a systematic analysis and visualization of how these weights influence the quality of generated counterfactual explanations Fig.\ref{fig:exp1_fig}.

Counterfactual explanations were generated for an input instance corresponding to a SEP event, targeting the opposite class label. By varying the weights across runs, we explored trade-offs among proximity, sparsity, diversity, and physics-guided constraints. The optimal configuration, determined using a weighted objective that balances proximity, sparsity, and diversity, was:
\text{proximity\_weight}$=$10, 
\text{sparsity\_weight}$=$0.1, 
\text{diversity\_weight}$=$1, 
\text{ordering\_penalty}$=$10; 

\subsection{Experiment 2: Comparative Study of Physics-Guided Counterfactual Explanations Against Standard DiCE}
\label{exp:exp2}
In this experiment, we evaluate and compare two counterfactual explanation methods-namely the standard approach implemented via a genetic DiCE explainer and the proposed physics-guided counterfactual generation framework (PGCE)-on a set of randomly selected test instances. For each instance, both methods are tasked with generating a fixed number of counterfactual explanations that flip the predicted class of a pre-trained Random Forest classifier. The generated counterfactual explanations are then quantitatively analyzed with respect to their proximity to the original instance, the sparsity of feature changes, the diversity among multiple counterfactual explanations, and the computational runtime.

Proximity is measured using the dynamic time warping (DTW)\cite{2007_Springer} distance, which accounts for temporal alignment in feature sequences, providing a robust measure of similarity between the original instance and its counterfactual explanations. Sparsity\cite{Mothilal2020} quantifies the fraction of unchanged features, reflecting how minimal the modifications are relative to the original input. Diversity\cite{Mazzine2021} captures the variation among the set of generated counterfactual explanations by computing the average pairwise Euclidean distance, thereby evaluating the framework’s ability to produce multiple plausible and distinct alternatives.

\begin{figure}[!h]
    \centering
    \includegraphics[width=\columnwidth]{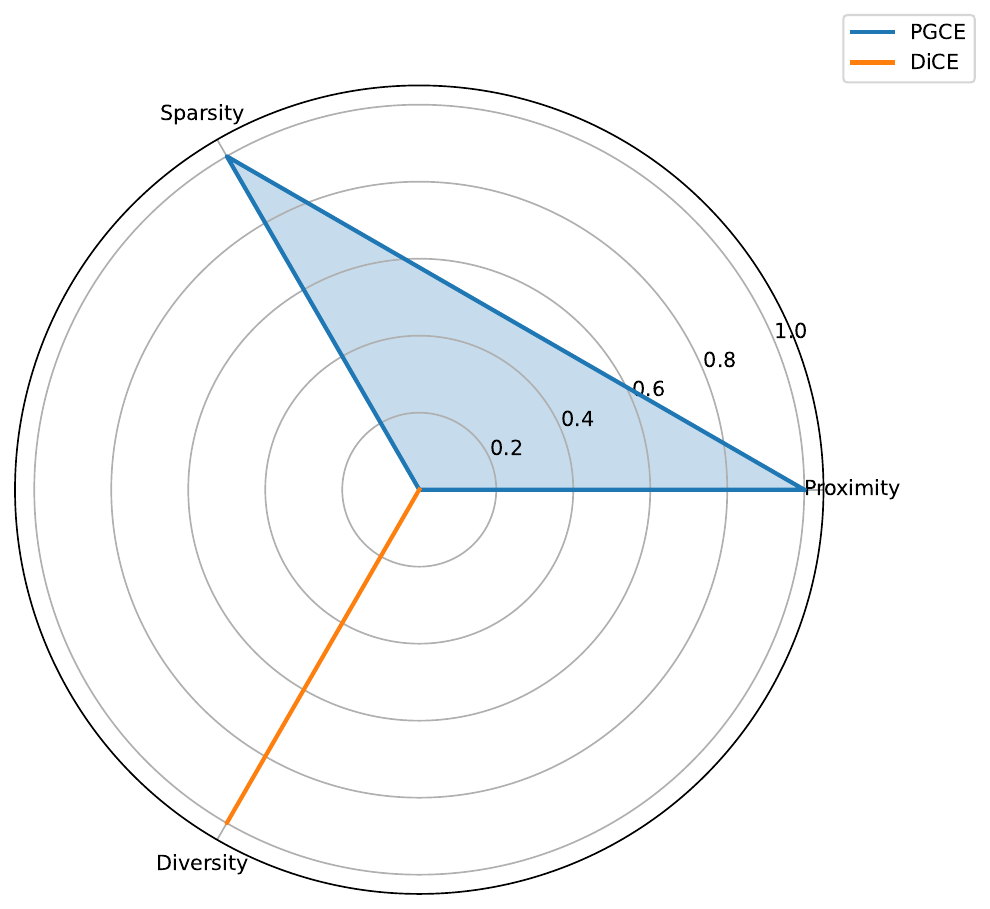}
    \caption{Comparison of Physics-Guided Counterfactual Explanations (PGCE) Framework and DiCE on Proximity, Sparsity, and Diversity (normalized to [0,1]).}
    \label{fig:radar_plot}
\end{figure}

The experiment is conducted in a loop over all selected instances, with each method generating a fixed number of counterfactual explanations per instance. Runtime is recorded for each generation process to assess computational efficiency. Finally, all metrics are aggregated across instances to provide a summary comparison between the two methods, highlighting their trade-offs in proximity, sparsity, diversity, and computational cost as displayed in Table \ref{tab:comparative_analysis}. To complement the tabular results, we visualize the comparative performance of both methods using a radar plot, as shown in Fig.~\ref{fig:radar_plot}. Each axis represents one of the key metrics-Proximity, Sparsity, and Diversity-normalized to the range [0,1] for comparability. 

\begin{table}[!h]
\caption{Comparison of Physics-Guided Counterfactual Explanations (PGCE) and DiCE on key metrics (mean $\pm$ std). ↑ indicates higher is better, ↓ indicates lower is better.}
\centering
\begin{tabular}{|c|c|c|c|c|}
\hline
Method & DTW ↓ & Sparsity ↑ & Diversity ↑ & Runtime ↓ \\
\hline
PGCE      & 4.42 $\pm$ 5.82 & 0.99 $\pm$ 0.01 & 5.49 $\pm$ 7.02 & 3.62 $\pm$ 0.29 \\
DiCE & 27.93 $\pm$ 8.89 & 0.97 $\pm$ 0.01 & 23.49 $\pm$ 4.40 & 6.99 $\pm$ 15.75 \\
\hline
\end{tabular}
\label{tab:comparative_analysis}
\end{table}

\subsection{Experiment 3: Fidelity Evaluation of Counterfactual}
\label{exp:exp3}
Counterfactual explanations were generated for test dataset instances corresponding to both SEP event and non-SEP event class labels. The fidelity of these counterfactual explanations was assessed using the approach described in \cite{Velmurugan2021}, where a pre-trained Random Forest classifier \cite{10386908} was employed to predict the class labels of the generated counterfactual explanations. Each generated counterfactual explanation was classified according to its intended target label, yielding a fidelity accuracy of 1.0 for both SEP and non-SEP instances. The associated classification metrics-including precision, recall, and F1-score were all 1.0 for both classes.

\section{Results and Discussions}
\label{sec:discussion}
\subsection{Hyperparameter Analysis}
\label{subsection:hyperparameter_analysis}
To understand the effect of hyperparameter choices on counterfactual quality, we conducted an experiment as describe in Section [\ref{exp:exp1}] over the proximity, sparsity, and diversity weights, along with the physics-guided ordering penalty introduced in Section [\ref{subsection:Problem_formulation}]. As illustrated in Fig.~\ref{fig:exp1_fig}, the hyperparameter analysis shows how weight settings influence the quality of counterfactual explanations, particularly in the context of large, high-dimensional multivariate time series data. Increasing the proximity weight encourages counterfactual explanations to remain close to the original instance, improving interpretability and minimizing unnecessary feature changes, which is critical when the data contains noise or missing values. Conversely, raising the diversity weight allows the exploration of alternative scenarios across correlated variables, but may reduce sparsity, producing explanations that modify more features. The sparsity weight and the physics-guided ordering penalty further modulate these trade-offs, ensuring counterfactual explanations remain minimally perturbed while adhering to domain-informed constraints, which helps maintain meaningful patterns in multivariate time series. The identified optimal configuration (proximity = 10, sparsity = 0.1, diversity = 1, ordering penalty = 10) achieves a balance between fidelity, sparsity, diversity, and physics-guided consistency, demonstrating the importance of careful hyperparameter tuning in the generation of reliable and interpretable counterfactual explanations for complex, high-dimensional datasets.

\subsection{Comparative Performance}
\label{subsection:Comparative_Performance}

The comparative analysis from Section [\ref{exp:exp2}] highlights how physics-guided counterfactual generation framework and DiCE navigate the trade-offs between proximity, sparsity, and diversity in counterfactual generation. Physics-guided counterfactual explanations' higher proximity and sparsity indicate that this framework produces explanations that are closely aligned with the original instances while modifying fewer features, which is particularly important in high-dimensional, multivariate time series data. In such datasets, preserving the temporal and cross-variable structure is critical to maintaining interpretability and avoiding unrealistic or infeasible counterfactual explanations. In contrast, DiCE achieves higher diversity, reflecting its capacity to explore a broader solution space and generate alternative scenarios. While this can be valuable for capturing different possibilities, it may also result in explanations that are less sparse and potentially harder to interpret, especially when multiple correlated features vary simultaneously.

These performance differences are further influenced by the nature and quality of the underlying data. Large-scale datasets often contain noise, missing values, or inconsistencies across variables, which can challenge generation of counterfactual explanations. By emphasizing proximity and sparsity, the method of generating physics-guided counterfactual explanations-mitigate the impact of such data quality issues, producing explanations that are robust and closely adhere to the original patterns in the multivariate time series data. DiCE’s broader diversity may amplify sensitivity to noise, highlighting the trade-off between exploring alternative explanations and maintaining reliability. Overall, the results demonstrate that incorporating physics-guided constraints allows the generation of counterfactual explanations that balance fidelity, minimal perturbation, and structural coherence across multiple variables, which is especially valuable for complex, high-dimensional, and noisy datasets.

\subsection{Fidelity Evaluation}
\label{subsection:Fidelity_Evaluation}
Finally, we evaluated the fidelity of the generated counterfactual explanations as described in Section [\ref{exp:exp3}] to ensure they align with the baseline model’s decision boundaries. Using the pre-trained Random Forest classifier \cite{10386908}, all counterfactual explanations generated from the testing dataset-including both SEP event and non-SEP event instances-were correctly classified as their intended target labels. This resulted in a fidelity accuracy of 1.0, with perfect precision, recall, and F1-score for both classes. These results confirm that the physics-guided counterfactual explanations framework-reliably produces counterfactual explanations that are both interpretable and faithful to the model’s predictions, validating the effectiveness and consistency of the framework across all test instances.

\subsection{Reconstruction Capability}
\label{subsection: Reconstruction_Capability}
The reconstruction capability introduced in Section [\ref{subsection:timeseries_reconstruction}] enables counterfactual explanations to be mapped back into physically realizable time series, preserving the temporal structure and correlations inherent in multivariate sequences. By iteratively adjusting slices of the original series according to the counterfactual values and averaging overlapping modifications as stated in Algorithm [~\ref{alg:generic_timeseries}], the physics-guided counterfactual explanations framework-produces perturbed signals that are both smooth and interpretable. This approach bridges the gap between conventional vector-based counterfactual explanations-which are limited in sequential domains and fully reconstructed time series.This allows domain experts to directly visualize how a counterfactual explanation scenario would manifest across the entire temporal signal.

The ability to generate realistic, temporally consistent signals is particularly important for high-dimensional, large-scale datasets, where traditional vector-based counterfactual explanations may obscure relationships between correlated features or fail to convey meaningful temporal patterns. By converting abstract feature perturbations into reconstructed signals, the framework facilitates big data visualization, enabling experts to assess counterfactual explanations in the same format as the original observations. This enhances interpretability, trust, and validation in domains such as SEP event prediction, biomedical monitoring, financial forecasting, and climate modeling, where understanding the full sequential context of counterfactual explanations is critical. In this way, reconstruction capability not only strengthens the explanatory power of physics-guided counterfactual explanations but also generalizes to a broad class of multivariate time series applications, providing a framework for interpretable and actionable counterfactual reasoning in sequential data.

\section{Conclusion and Future Scope}
This paper presented Physics-Guided Counterfactual Explanations, an application-driven extension of existing counterfactual frameworks tailored for time-series classification under scientific constraints. While counterfactual generation has been studied extensively in tabular and image domains, our novelty lies in embedding physics-based feasibility constraints into the generation process, ensuring that counterfactual explanations remain consistent with underlying system dynamics. Applied to solar energetic particle (SEP) forecasting, this framework achieves significantly lower DTW distance, higher sparsity, and reduced runtime compared to DiCE. These results show that even as an adaptation of an existing approach, physics-guided counterfactual explanations introduce a new perspective by aligning counterfactual explanations with physical plausibility, making them more actionable and trustworthy in scientific domains.

Looking ahead, this framework provides a foundation for interpretable AI methods in large-scale and big data environments, where the ability to generate scientifically valid counterfactual explanations at scale is increasingly important. Future work will explore optimization strategies for scalable counterfactual generation, such as distributed computation and streaming adaptations for continuous data flows. Another promising direction is the integration of physics-guided counterfactual explanations with high-capacity forecasting models (e.g., deep learning architectures) while preserving interpretability. These directions highlight the broader potential of physics-guided counterfactual explanation framework as a bridge between machine learning interpretability and domain knowledge in the era of big data.

\section{Acknowledgment}
This work is supported in part by two National Science Foundation (NSF) awards (\#2104004 and \#2513886).

\section*{GenAI Usage Disclosure}
Generative AI tools were used exclusively for grammar correction and language editing during manuscript preparation. All research, code and data analysis was performed by the authors.

\sloppy
\bibliographystyle{IEEEtran}
\bibliography{IEEEabrv,ref}

\end{document}